\documentclass[runningheads]{llncs}
\usepackage{graphicx}

\usepackage{subfig}
\usepackage{tabularx}
\usepackage{tabulary}
\usepackage{caption}

\begin{document}
\title{Mitigating Gender Bias in Machine Learning Data Sets}
%
\titlerunning{Mitigating Gender Bias in Machine Learning Data Sets}
%
\author{ Susan Leavy \and Gerardine Meaney \and Karen Wade \and Derek Greene }
\institute{University College Dublin, Ireland\\
\email{\{susan.leavy,gerardine.meaney,karen.wade,derek.greene\}@ucd.ie}}
\authorrunning{S. Leavy et al.}
\maketitle              

 
\begin{abstract}
Algorithmic bias has the capacity to amplify and perpetuate societal bias, and presents profound ethical implications for society. Gender bias in algorithms has been identified in the context of employment advertising and recruitment tools, due to their reliance on underlying language processing and recommendation algorithms. Attempts to address such issues have involved testing learned associations, integrating concepts of fairness to machine learning, and performing more rigorous analysis of training data. Mitigating bias when algorithms are trained on textual data is particularly challenging given the complex way gender ideology is embedded in language. This paper proposes a framework for the identification of gender bias in training data for machine learning. The work draws upon gender theory and sociolinguistics to systematically indicate levels of bias in textual training data and associated neural word embedding models, thus highlighting pathways for both removing bias from training data and critically assessing its impact in the context of search and recommender systems.



\keywords{algorithmic bias \and gender \and  machine learning \and natural language processing.}
\end{abstract}


\section{Introduction}
Algorithmic bias, as embedded in search and recommendation systems, has the capacity to profoundly influence society. For instance, recommendation systems targeting employment-related advertisements were found to demonstrate gender bias \cite{lambrecht2019algorithmic}. The gendering of personal assistant technologies as female is also being questioned as constituting indirect discrimination, potentially contravening international women`s rights law \cite{adams2019addressing}.  With the rise in the use of facial recognition in areas such as border control, along with the issues with variance in accuracy depending on gender and race \cite{buolamwini2018gender}, there is a risk that bias will be incorporated directly into the core public infrastructure of a country. Even legal systems are vulnerable to the influence of algorithmic bias through the use of systems such as \emph{Compas}, where recommendations around parole lengths have demonstrated evidence of racial bias \cite{angwin2016machine}.


The source of this kind of bias often lies in the way societal inequalities and latent discriminatory attitudes are captured in the data from which algorithms learn.  Given the ways in which sentiments regarding race and gender ideology can be deeply embedded in natural language, uncovering and preventing bias in systems trained on such unstructured text can be particularly difficult. This paper focuses on algorithmic gender bias, and proposes a framework whereby language based data may be systematically evaluated to assess levels of gender bias prevalent in training data for machine learning systems. The framework is developed by accessing potential bias prevalent in articles in a popular UK mainstream media outlet, \textit{The Guardian}, over a decade from 2009 to 2018. This is contrasted with biases uncovered in a corpus of 16,426 digitised volumes of 19th-century fiction from the British Library. This paper demonstrates how bridging AI and research in gender and language can provide a framework for potentially gender-proofing AI, and contributes to ongoing work on the systematic mitigation of algorithmic gender bias.

\section{Related Work}
Strategies to test for algorithmic gender bias have involved evaluation of system accuracy and learned associations in machine learning technologies that underlie many search and recommendation systems \cite{dixon2018measuring}.  Implicit Association Tests (IATs) were found to be effective in uncovering gender bias in the `common crawl' corpus, a large collection of text sourced from the web \cite{caliskan2017semantics}. Stereotypical representations of gender were also identified in an analysis of an embedding model trained on Google News content \cite{bolukbasi2016man}.  Evidence of 100 years of gender bias in relation to employment and associated adjectives was uncovered by applying word embedding techniques to text sourced from the Corpus of Historical American English, Google Books, New York Times, and Google News \cite{garg2018word}. 
The introduction of concepts of fairness to machine learning and modifying learned associations in algorithms have been used to address gender bias \cite{zhang2018mitigating}. Disassociating biased relationships between entities in word embedding models has reduced stereotypical associations between, for instance, gender and types of employment \cite{bolukbasi2016man}. However, studies have shown that implicit gender bias persists despite these de-biasing methods \cite{gonen2019lipstick}. The modification of training corpora prior to learning of gender bias has been explored through the provision of training data where the gender of entities in the corpora are swapped and has been proven to reduce gender bias in predictions \cite{zhao2018gender}. Building on these approaches, this paper explores the opportunity to incorporate findings from research in the gender theory and feminist linguistics which has sought to uncover the features of language that encode gender bias, in order to develop scalable methods to systematically identify bias in training data. 

\subsection{Uncovering Gender Bias}


The crucial influence of language in shaping and reinforcing gender in society is explored within the field of feminist linguistics identifying language features that encode bias \cite{mills1995feminist}. For instance, premodified terms such as `female lawyer' or `female police officer', are interpreted as highlighting their existence as contrary to societal expectations \cite{sigley2002looking}.  Similarly, terms such as 'career woman' or 'working mother' don not have popular equivalents for men \cite{romaine1998communicating}. How language change reflects underlying changes in prevalent gender ideology in society is demonstrated by the increasing use of `they/them' rather than `he/him' and `humanity' rather than `mankind', and the replacement of `Mrs' and `Miss' with `Ms' \cite{baker2008sexed}. Such shifts in language use indicate the potential for language corpora to preserve and potentially perpetuate outdated concepts of gender. 

Of particular relevance in the context of AI-supported recommender systems ans web search is the tendency shown in the media to refer to adult women as  `girls' \cite{sigley2002looking}. Women have also been shown to be more associated with derogatory, sexual and negative descriptions\cite{baker2008sexed,caldas2010curvy,pearce2008investigating}. Associations between women, beauty and lack of agency have also been identified as encoding gender bias \cite{frith2005construction,mills1995feminist}. 

Measurements of the presence of women in text has shown to be an effective measure of potential gender bias \cite{ali2010automating,shor2014time}. More subtle measures of potential gender bias could also be considered. For instance, conventions regarding how binomials are ordered in English dictates that the most powerful is named first (e.g. doctor/nurse, teacher/pupil). However, gender is the most important determiner of order, thus revealing a concept of social order assigning more power to men \cite{mollin2012revisiting,vefali2010coordinate,motschenbacher2013gentlemen}. 

In devising methods to identify gender bias in algorithms, studies have incorporated researchers' or crowd-sourced interpretations of what constitutes gender stereotypes \cite{bolukbasi2016man,garg2018word,swinger2019biases}. Building on this, this paper proposes a framework whereby language-based training data may be systematically gender-proofed to mitigate bias in machine learning algorithms.

\section{Methods}

Given that early studies of bias in the representation of women focused study of literature, we analyse a set of over 16,000 volumes of 19th-century fiction from the British Library Digital corpus \cite{leavy19curatr}. This corpus was selected due to the well-documented evidence of stereotypical and binary concepts of gender in 19th-century fiction \cite{ingham2002language}, and therefore represents a useful source of baseline data, allowing methods to be tested and refined, and subsequently generalised to other corpora. To investigate evidence of gender bias in contemporary corpora, this research analyses a decade of articles from the UK newspaper, \textit{The Guardian} including every article published online between 2009 and 2018, as retrieved from The Guardian Open Platform API\footnote{\url{https://open-platform.theguardian.com}}.

\emph{Word embeddings} refer to a family of machine learning approaches that yield numeric, low-dimensional representations of words based on lexical co-occurrences. We focus on these models in our work, as they are widely used as a building block for further downstream analysis in many language processing tasks \cite{tang2014learning}. These approaches have also been successfully used to uncover patterns of stereotypical gender-based associations \cite{bolukbasi2016man,caliskan2017semantics,garg2018word}. Following these approaches, we investigate conceptual relationships in the texts using embedding representations. The conceptual relationships examined for evidence of gender bias were informed by a framework based on feminist critiques and analysis of the use of language. This framework focused on linguistic features that encode gender bias, and was used to inform both the development of thematic lexicons and the selection of features from the corpora, specifically:
\begin{itemize}
  \item Presence of women in text
  \item Gender-specific terms (e.g. career woman)
  \item Premodified terms (e.g. female lawyer)
  \item Androcentric terms and misuse of gender neutrals
  \item Negative or stereotypical associations
\end{itemize}

The particular word embedding variant used in this work is a 100-dimensional Continuous Bag-Of-Words (CBOW) \emph{word2vec} model \cite{mikolov2013efficient}, trained on the full-text volumes of the 16,426 fictional texts from the British Library corpus. \emph{Word lexicons} can used to represent concepts of gender and themes related to bias. In our work, lexicons are constructed by defining an initial small set of seed terms, and expanding this set using related words as determined by similarities derived from the embedding model. Contemporary thematic lexicons which were used to examine gendered associations within the text were based on\textit{ The General Inquirer} dictionaries\footnote{\url{http://www.wjh.harvard.edu/inquirer}}. Given the consistent findings within gender theory of the portrayal of women in texts as passive, emotional and defined in the context of family relationships, the themes focused on involved the General Inquirer semantic categories pertaining to emotion, family and terms that convey activity. The semantic category pertaining to moral judgement and misfortune (vice) is also explored to capture an idealised concept of femininity that is evident in Victorian literature and examine changes within contemporary culture. 

The relationships between conceptual lexicons in the corpus were visually explored using the Tensorboard tool\footnote{\url{https://www.tensorflow.org}}. Relational patterns were then analysed by calculating cosine distances between terms within the embedding model. These were depicted visually to highlight differences in how terms in lexicons representing gender were related with other concepts in the text. Rule-based information extraction was also used to evaluate the volume of representations of men and women in text and to extract particular linguistic features, such as the ordering of binomials.


\section{Findings and Analysis}
This research demonstrates an approach for developing metrics for bias in data sets informed by feminist linguistics and gender theory, in order to mitigate algorithmic bias. We see that gender bias was uncovered in neural word embedding models trained on both historical and contemporary data-sets thus presenting scalable techniques for automatically assessing data sets for evidence of bias.

\subsection{Presence of Women in Text}

\begin{figure}[!b]
  \centering
  \subfloat[Guardian and 19th c. corpora.]{\includegraphics[width=0.5\textwidth]{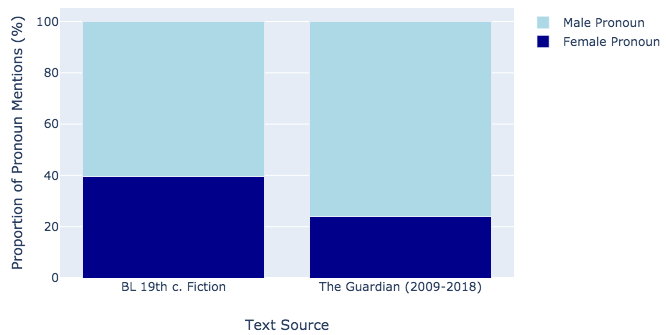}\label{heshe_total}}
  \hfill
  \subfloat[Guardian (2009-2018).]{\includegraphics[width=0.5\textwidth]{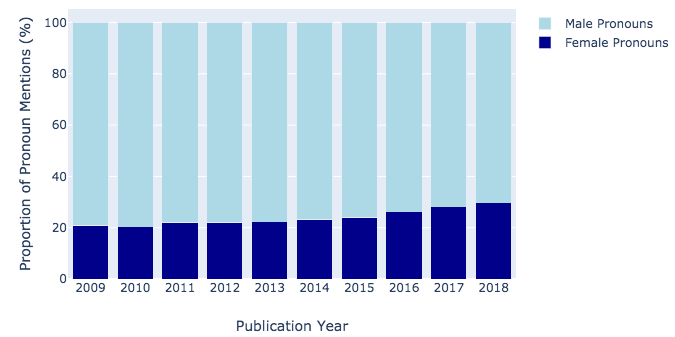}\label{heshe}}
  \caption{Presence in of women in text, as reflected by pronoun usage in The Guardian and 19th-century British fiction corpora.}
\end{figure}

The presence of women in data sets is a simple but highly effective metric of bias 
in the Guardian as measured by the proportional occurrence of male and female pronouns was distinctly lower than that in the corpus of 19th-century fiction (Fig. \ref{heshe_total}). While a higher representation of women is arguably to be expected in the 19th-century volumes, it is also lower also than an analysis of the New York Times which found female representation of 28\% in 2008 \cite{shor2014time}. Only 20\% of gendered pronouns in the year following that in The Guardian were female. However, there has been a steady increase to 30\%  female representation in The Guardian by 2018 (Fig. \ref{heshe}).  Based on an evaluation of gender bias by the metric of volume of coverage alone, The Guardian appears to be more biased than 19th-century British fiction, pointing towards the need for further semantic analysis of the texts.

\subsection{Gender-Specific Terms}

The premodification of terms can introduce a gender dimension to concepts that can often convey stereotypes and imply information about gender that is biased. In the 19th-century for example, there was a prevailing idealised concept of femininity that saw certain attributes as distinctly female (e.g. female nature). This is reflected by the fact that the term `female' appears 2.5 times more frequently than the term `male'. This also points towards `male' being considered the default in many contexts, and `female' the exception that should be named (see. \cite{perez2019invisible}). Following this rationale, the lowering of proportional use of the term `female' to 56\% in 2009 suggests a lessening of gender bias. However, this figure increases to 60\% in 2018, potentially due to a greater level of gender discourse in the media during this year, demonstrating the importance of take context into account when attributing gender bias to a particular collection of texts. 

\subsubsection{Gender-specific occupations.}
The context of gender premodification was analysed and classified according to those pertaining to occupations, characteristics and references to the physical body. The volume of terms related to occupations that are specified by gender notably increased by the end of the decade from 2009 in The Guardian. This increase is not, however, exclusive to women, and demonstrates a potential new dimension in the analysis of gender bias in language. In 19th-century fiction, male premodified occupations were rare and the three examples found referred to roles that both men and women undertook (see Table \ref{predmodified_occupations}). 

In 2009 in The Guardian, occupations specified as male were primarily related to occupations that were often shared or roles predominantly held by women. For example, `nurse' is primarily a female occupation, so a male nurse is identified, through premodification, as an exception. However, by 2018 there is a dramatic increase in premodified occupations that are stereotypically male. For example, `doctors', `footballers', `executive' are premodified as male in 2018. A similar increase is evident in the use of terms that are specified as female. Overall, however, occupations that are conceptually associated with both genders equally, denoted by the terms being premodified equally by both genders (e.g. `writer', `journalist'), remain a small proportion of the gender-specific terms that were used. A potential cause for the increase in the occupations specified by gender may be media discussion of workplace equality. Therefore, a calculation of gender specified occupations may not reflect gender bias, but the presence of feminist discourse arguing for gender equality. These finding suggest that a more reliable measure of gender equality is the number of occupations that are equally premodified by gender, where neither is considered the default gender for a given role.


\subsubsection{Gender-specific characteristics.}

By extracting characteristics that are specified as female from the British Library corpus, we captured the Victorian associations of women with `loveliness', `weakness' and `modesty' (see Table \ref{predmodified_character}). This contrasts with female `empowerment', `power', and `talent' in 2009 in The Guardian. However, those associated with men in 2009 reflect stereotypical concepts of violence and dominance. There was a striking increase in the use of premodified characteristics by 2018 with the introduction of terms that echo feminist discourse. 

These findings demonstrate that, even though mentions of gendered characteristics in relation to men and women may occur in the context of articles critiquing stereotypes, depending on the application of a machine learning algorithm, these associations may still be learned and may perpetuate the very stereotypes the articles propose to disrupt. For instance, the association in the 2018 Guardian data between female `hysteria' and `fragility' and male `privilege' might not reflect bias on the part of the authors, yet uncovering these associations systematically demonstrates how gendered character traits could be learned by a machine learning algorithm from such a training corpus. 

\subsubsection{Gender-specific physical terms.}
The corpus of 19th-century fiction, as expected, reflects abstract and potentially metaphorical references to gender-specific physical aspects of the human body (Table \ref{predmodified_body}). In The Guardian corpus these descriptions are more direct. However, in 2018 there is a notable increase in the number of terms premodified by both `male' and `female'. This further supports the proposal suggested in relation to occupations, that a solid indicator of bias may be a relatively higher rate of terms that are equally premodified for men and women.

\begin{table*}[ht]
\scriptsize
\caption{Premodified occupations in order of frequency}
\vskip 0.6em
\begin{tabularx}{350pt}{c X X}
Corpus & Male Premodified & Female Premodified \\ \hline

    \emph{British Library} & \textbf{servant}(s) \textbf{domestic}(s) \textbf{attendant}(s) & \textbf{servant}(s) \textbf{attendant}(s) warrior(s) \textbf{domestic}(s) slave(s) art(ist(s)), novelist(s) detective(s) sovereign(s) warder labour missionary(ies) singers(ing) highwayman writers employment teacher(s) philosopher doctor poets assistant forger students cook politician industry occupation proprietor warders \textbf{(28 unique terms)}\\ \\
    \emph{Guardian 2018} &  \textbf{writers} \textbf{artists} \textbf{actors} players \textbf{employees} \textbf{artist} \textbf{authors} \textbf{writer} \textbf{mps} actor player athletes \textbf{directors} \textbf{models} \textbf{politicians} \textbf{stars} presenters critics model \textbf{journalists} \textbf{officers} director doctors \textbf{dancer} \textbf{dancers} \textbf{staff} co-stars footballers athlete football \textbf{officer} author executives \textbf{teacher} applicants \textbf{celebrities} comedians journalist musicians novelist scientists \textbf{star} \textbf{workers} academics \textbf{boss} comic doctor investors \textbf{police} presenter \textbf{teachers} \textbf{bosses} ... \textbf{(145 unique terms)}
 & \textbf{artists} \textbf{staff} \textbf{directors} \textbf{candidates} \textbf{employees} students \textbf{writers} \textbf{artist} athletes director \textbf{politicians} \textbf{workers} president \textbf{journalists} governor film-makers footballers doctors \textbf{authors} doctor \textbf{stars} musicians chief presenters scientists \textbf{writer} composers \textbf{police} \textbf{teachers} coaches employee jockeys singer \textbf{officer} mayor candidate journalist performers pilots student comics singers entrepreneurs \textbf{officers} cast jockey reporter athlete chef chefs engineers \textbf{politician} senator ... \textbf{(263 unique terms)}\\\hline
\end{tabularx}
\label{predmodified_occupations}
\end{table*}

\begin{table*}[ht]
\scriptsize
\caption{Premodified characteristics}
\vskip 0.6em
\begin{tabularx}{350pt}{c X X}
Corpus & Male Premodified & Female Premodified \\ \hline

    \emph{British Library} & violent(ce), mind, \textbf{character(s)}, \textbf{young}, beauty, \textbf{intellect}, youth \textbf{(7 unique terms) }& heart(s), \textbf{character(s)}, mind(s), loveliness, education, influence, nature, charms, virtue, curiosity, vanity, ailments, delicacy, excellence, \textbf{intellect}, heroism, \textbf{young}, instinct, taste, innocence, soul, purity, propriety, grace, perfection, weakness, affection, finesse, modesty, ingenuity, monster, sympathy, tactics, errors, old, pride, dignity, honour, spirit \textbf{(40 unique terms)} \\
    
    \emph{Guardian 2009} &  voice bonding dominated attention \textbf{characters} voices grooming dominance violence \textbf{character} domination primary \textbf{brain} ego gaze heroes \textbf{power} behaviour life preserve bravado chauvinist elite performance privilege rage \textbf{(26 unique terms) }& \textbf{characters} \textbf{character} talent empowerment emancipation perspective adolescence \textbf{power} soul stereotypes action acts \textbf{brain} \textbf{(14 unique terms)}
\\
 
\emph{Guardian 2018} & \textbf{gaze} \textbf{characters} privilege sexual lead \textbf{voice} dominance voices entitlement dominated \textbf{behaviour} character supremacy perspective \textbf{desire} identity rage winner culture fantasy leaders mental performance pleasure pride aged ego genius leads authority literary psyche aggression misbehaviour perpetrators political problem energy environment life queerness anxiety approval attitudes chauvinism chauvinist domination fantasies glance grooming ... \textbf{(123 unique terms)}

& \textbf{characters} representation empowerment character voices experience \textbf{voice} \textbf{gaze} aged power solidarity identity \textbf{desire} agency narrator autonomy superhero ambition presence artistic representatives strength senior state anger \textbf{behaviour} liberal narratives women achievement brain creative energy equality imagination objectification resistance social wits brains creativity fantasy freedom friendly gender hereditary independence love relationship ... \textbf{(92 unique terms)}\\
\hline
\end{tabularx}
\label{predmodified_character}
\end{table*}

\begin{table*}[!t]
\scriptsize
\caption{Premodified physical references.}
\vskip 0.6em
\begin{tabularx}{350pt}{c X X}
Corpus & Male Premodified & Female Premodified \\ \hline

    \emph{British Library} & \textbf{figure}(s) \textbf{eye(s)} \textbf{sex} heart \textbf{head} \textbf{hand}
 & \textbf{figure}, form(s), \textbf{sex}, beauty, \textbf{hand(s)}, attire(d), \textbf{head(s)}, face(s), \textbf{eye(s)}, breast, shape, lips, tongue(s), bosom(s), flesh \textbf{(16 unique terms)}\\
    \emph{Guardian 2009} & beauty \textbf{sex} \textbf{genitalia} \textbf{sexual} figures body fertility figure hormone psyche \textbf{(10 unique terms)} & sexuality genital \textbf{sexual} body form \textbf{genitalia} \textbf{sex} beauty face anatomy faces figure vocals orgasm \textbf{(14 unique terms)}\\
    \emph{Guardian 2018} & \textbf{body} \textbf{infertility} \textbf{sex} \textbf{bodies} \textbf{suicide} \textbf{fertility} \textbf{figure} \textbf{genitalia} \textbf{beauty} \textbf{figures} \textbf{form} gender makeup face clothes clothing eyes faces hormone hormones orgasm sperm anatomy flesh gay hair health hormonal libido physique reproduction reproductive suicides \textbf{(33 unique terms)}
 & genital \textbf{body} sexuality \textbf{form} sexual \textbf{beauty} orgasm \textbf{sex} \textbf{bodies} \textbf{figure} \textbf{genitalia} pain \textbf{figures} flesh reproductive anatomy biology face masturbation genitals same-sex \textbf{suicide} \textbf{fertility} gay hormones cancers faces health breast contraceptive hormone \textbf{infertility} nipples bodily breasts orgasms pregnant skin sterilisation\textbf{ (39 unique terms)}\\
  \hline
  \end{tabularx}
  \label{predmodified_body}
\end{table*}

\subsection{Trends in use of Androcentric Generics and Gender Neutrals}

The term `mankind' is often used as a gender-neutral term. However, research dating back to the 1970's demonstrates that such terms are not perceived as inclusive \cite{martyna1978does}. As expected, androcentric gender neutrals were  commonplace in 19th-century but also appears surprisingly often. The use of gender-neutral terms such as `chairperson' and `statesperson' is negligible. While the proportion of female MPs in the UK is 30\%, the fact that the gender neutral term `statesperson' is not applied to them but `statesman' is commonly used, suggests that the role remains conceptually male. The use of contemporary gender-neutral terms therefore would indicate levels of gender bias in a corpus.




\subsection{Gendered Associations: Negative or Stereotypical Descriptions}

Conceptual associations between gender and particular themes were assessed with neural word embedding. Conceptual lexicons based on the General Inquirer that were analysed included emotion, terms denoting family, action and vice (described as an assessment of misfortune or moral disapproval). 

\subsection{Gender and Emotion}
\begin{figure}[!t]
\centering
\includegraphics[width=.7\linewidth]{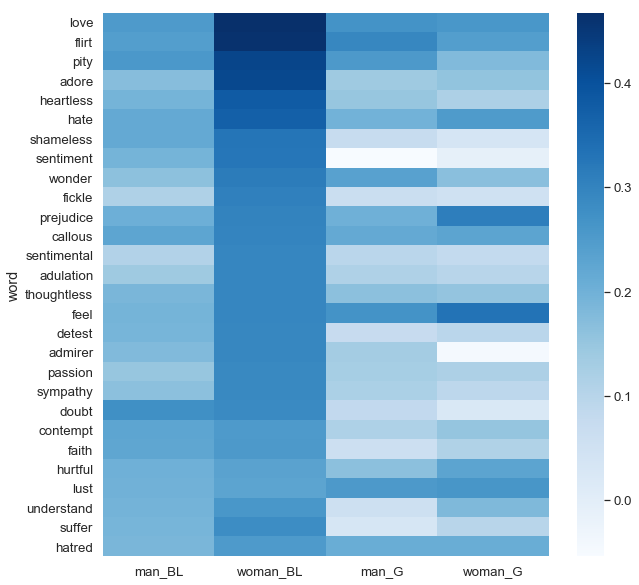}
\caption{Emotion: Similarity of top terms for the BL and The Guardian corpora.}
\label{emotion_GBL}
\end{figure}

The analysis of cosine similarity of terms within the word embeddings uncovered distinctly stereotypical associations of gender and emotion for the BL corpus, as we might expect from 19th-century fiction. The top 20 terms denoting emotion associated with men and women were extracted and the levels of association for both the historical and contemporary corpora presented in Figure \ref{emotion_GBL}. Overall, women were associated with emotion substantially more than men (`women' with 0.101 vs. `men' 0.056 mean cosine similarity). In contrast, in The Guardian corpus the overall association of men and women with terms denoting emotion was almost equal (`women' 0.078 vs. `men' 0.089 mean cosine similarity). 

\subsection{Gendered Action}

\begin{figure}[!t]
\centering
\includegraphics[width=.7\linewidth]{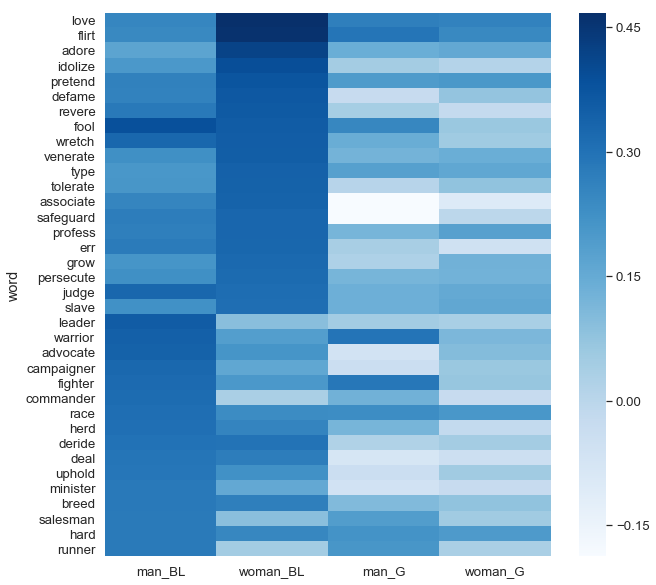}
\caption{Action: Similarity of top terms for the BL and The Guardian corpora.}
\label{action_GBL}
\end{figure}

The association of terms denoting action in the corpus of 19th century support the theory that men were portrayed in more active and women in more passive terms (Fig. \ref{action_GBL}. Men are most closely associated with terms including `leader', `warrior', `advocate', `campaigner`, `fighter`, and `commander'. This contrasts distinctly with the kinds of actions women were associated with, including `love', `flirt', `adore', `idolize', and `pretend'. These distinctive associations did not continue in The Guardian corpora, but present more subtle differences and reflect contemporary issues, as indicated by the level of co-occurrence of terms like `harass' and `liberation' with `women' in 2018 (Table \ref{Activitychanges}).

\begin{table}[ht]
\scriptsize
\caption{Action lexicon: Gendered associations in The Guardian corpora.}
\begin{tabular}{p{1cm}p{11cm}}
\hline
\multicolumn{2}{c}{Female}\\ \hline		
2009	& intercourse divorce groom nurse molest dress violence skin wear cuddle driver participant drink obedient articulate actor abuse antagonistic seeker murder\\
	    
2018	&  representation intercourse abuse actor violence skin speak wear liberation assault articulate driver nurse dress aspire violent humiliate harass behavior  \\
\hline

\multicolumn{2}{c}{Male}\\\hline
2009	& driver stab boxer killer cuddle groom hug love occasion nurse lying guard actor compliment fan stroke wear crowd murder stood \\

2018	& driver compliment warrior figure fuck saw stab alive humiliate fan actor boxer guess killer reason occasion wear gone motivation \\  \hline
\end{tabular}

  \label{Activitychanges}
\end{table}

\subsection{Character Descriptions and Gender}

\begin{figure}[!t]
\centering
\includegraphics[width=.7\linewidth]{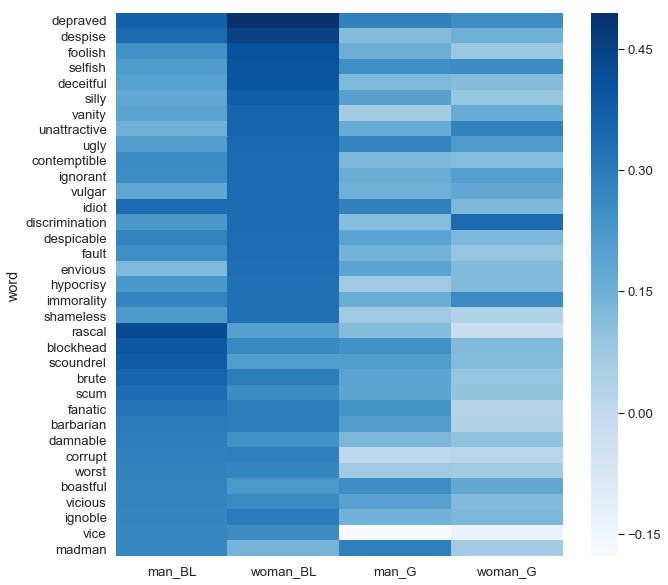}
\caption{Vice: Similarity of top terms for the BL and The Guardian corpora.}
\label{vice_GBL}
\end{figure}

The concept of vice for women in the 19th-century was particularly gendered, and this is reflected in the top terms from the General Inquirer lexicon that are associated with women in the corpus of British fiction (Fig. \ref{vice_GBL}). Here we see that women are most associated with terms referring to silliness and moral failings. What is unexpected, however, is that among all the themes, the levels of association of individual words seems to have remained the most consistent. 

The terms relating to concepts of vice that are associated with men and women in the The Guardian reflect distinct patterns (Table \ref{Vicechanges}). Those associated with women echo contemporary media discourse on sexual violence. While terms pertaining to relationships, including `divorce', `unfaithful', and `adultery', are associated with women, there are no equivalents associated with men.  Terms denoting vice associated with men largely pertain to judgements of character (e.g. `drunk', `crazy', `selfish', `madman', `idiot', `arrogant', `cruel', `stupid').

\begin{table}[!t]
\scriptsize
\caption{Vice lexicon: Gendered associations in The Guardian corpora.}
\begin{tabular}{p{1cm}p{11cm}}
\hline
\multicolumn{2}{c}{Female}\\	
\hline
2009	& divorce discrimination loveless adultery drunk insecure indecent violence unfaithful stigma suicide sick cruel illness depraved selfish vile ignorant abuse\\
	    
2018	& stigma discrimination abuse trauma violence insecure suicide sick inferior depression adultery assault blindness ordeal unjust coercion violent unsure condescending vulnerable\\
\hline

\multicolumn{2}{c}{Male}\\\hline
2009	& drunk misfortune ordeal vain idiot arrogant cruel stupid vile mad naive forgetfulness damned foolish ugly unbelievable awful loveless fanatic murder\\

2018	& drunk crazy selfish madman rascal horrible arrogant stupid suicide idiotic inferior foolish audacity idiot ungrateful guilty assault adversity unlucky badly \\  \hline
\end{tabular}

  \label{Vicechanges}
\end{table}

\subsection{Gendered Associations with Family}

\begin{figure}[!t]
 \centering
    \includegraphics[width=.7\linewidth]{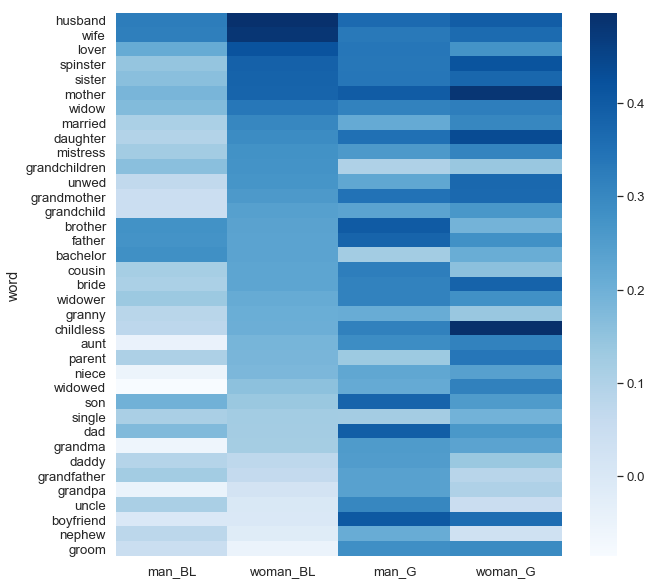}
 \caption{Family: Similarity of top terms from BL Corpus with The Guardian.}
  \label{BL_G_Family}
\end{figure}

Gender bias is evident in the gendered associations present in the neural word embedding model pertaining to 19th-century fiction with terms denoting family. Men in this corpus had little association with concepts of family, when compared to women (see Fig. \ref{BL_G_Family}). Evidence suggests that this has changed in contemporary culture, with overall associations appearing equal. However, women are distinctly more frequently associated with the status of parenting, as `mother' or `childless'.

\subsection{Ordering of Binomials}

Women were listed after men in examples of gendered binomials in 87\% of cases appearing in the corpus of 19th-century British fiction. The cases analysed involved listings of wife, husband, girl, boy, son, daughter, man, women, men, women. Listings were captured using a rule-based extraction process where excerpts containing both terms were identified and evaluated. In The Guardian news articles, this occurred 78\% in 2009, dropping to 74\% in 2018. Listing husbands before wives was the most persistent case, remaining at 87\% and 84\% respectively for the 2018 and 2009  collections, suggesting the concept of marriage is most closely tied with power relations.  This finding of a relationship between power, gender, and the ordering of binomials suggests that augmenting ordering in training data may prevent the learning of underlying structures in language denoting a societal conception of the most powerful.

 \section{Conclusion}

The findings of this research demonstrate how methods from machine learning, used within a framework informed by feminist linguistics and gender theory, can be used to evaluate levels of gender bias within natural language training corpora. A corpus of 19th century fiction along with a contemporary data set comprising every article published online in The Guardian newspaper over the decade between 2009 and 2018 was examined. The methods developed in this research uncovered gendered patterns in the corpus of 19th-century fiction that reflected Victorian concepts of gender while analysis of The Guardian uncovered linguistic patterns that capture contemporary concepts of gender. The emergence of feminist discourse in the media is also evident through gendered associations captured in word embedding uncovering an intriguing finding concerning how critiques of gender stereotypes could in fact generate stereotypical associations in neural embedding model. The systematic approach for capturing gender bias outlined in this paper is scalable and may be applied to a broad range of corpora, presenting new pathways for automatically assessing levels of bias in training corpora for search and information extraction systems.

\vskip 2em
\section*{Acknowledgements} This research project was supported by the Irish Research Council (IRC) and Science Foundation Ireland (SFI) under Grant Number SFI/12/RC/2289\_P2.

%
%
\bibliographystyle{splncs04}

\end{document}